\documentclass{article}
\pdfoutput=1
\usepackage{ijcai21}

\usepackage{amsmath,amsfonts,bm}

\def\eqref#1{equation~\ref{#1}}

\def\1{\bm{1}}

\DeclareMathAlphabet{\mathsfit}{\encodingdefault}{\sfdefault}{m}{sl}
\SetMathAlphabet{\mathsfit}{bold}{\encodingdefault}{\sfdefault}{bx}{n}

\DeclareMathOperator*{\argmax}{arg\,max}

\usepackage{times}

\usepackage{soul}
\usepackage{url}
\usepackage[hidelinks]{hyperref}
\usepackage[utf8]{inputenc}
\usepackage[small]{caption}
\usepackage{graphicx}
\usepackage{amsmath}
\usepackage{booktabs}
\usepackage{xcolor}
\usepackage{caption}
\usepackage{subcaption}
\usepackage{multirow}

\usepackage{wrapfig}

\newcommand*{\img}[1]{%
    \raisebox{-.2\baselineskip}{%
        \includegraphics[
        height=0.9\baselineskip,
        width=0.9\baselineskip,
        keepaspectratio,
        ]{#1}%
    }%
  }

\newtheorem{example}{Example}

\newtheorem{definition}{Definition}[section]

\title{Abductive Knowledge Induction From Raw Data}

\author{
Wang-Zhou Dai\and
Stephen H. Muggleton\\
\affiliations
Department of Computing, Imperial College London, London, UK\\
\emails
\{w.dai, s.muggleton\}@imperial.ac.uk
}

\begin{document}

\maketitle

\begin{abstract}
For many reasoning-heavy tasks involving raw inputs, it is challenging to design
an appropriate end-to-end learning pipeline.
Neuro-Symbolic Learning, divide the process into sub-symbolic perception and
symbolic reasoning, trying to utilise data-driven machine learning and
knowledge-driven reasoning simultaneously. However, they suffer from the
exponential computational complexity within the interface between these two
components, where the sub-symbolic learning model
lacks direct supervision, and the symbolic model lacks accurate input facts.
Hence, most of them assume the existence of a strong
symbolic knowledge base and only learn the perception model while avoiding a
crucial problem: where does
the knowledge come from? In this paper, we present Abductive
Meta-Interpretive Learning ($Meta_{Abd}$) that unites abduction and induction
to learn neural networks and induce logic theories jointly from
raw data. Experimental results demonstrate that $Meta_{Abd}$ not only
outperforms the compared systems in predictive accuracy and data efficiency but
also induces logic programs that can be re-used as background knowledge in subsequent
learning tasks. To the best of our knowledge, $Meta_{Abd}$ is the first system
that can jointly learn neural networks from scratch and induce recursive
first-order logic theories with predicate invention.
\end{abstract}

\section{Introduction}

Despite the success of data-driven end-to-end deep learning in many traditional
machine learning tasks, it has been shown that incorporating domain knowledge is
still necessary for some complex learning
problems~\cite{dhingra:differentiable,grover:neuralsort,nalu18}.
In order to leverage complex domain knowledge that is discrete and relational,
end-to-end learning systems need to represent it with a differentiable module
that can be embedded in the deep learning context. For example, graph neural
networks (GNN) use relational graphs as an external knowledge
base~\cite{zhou18:gnnsurvey}; some works even considers more specific domain
knowledge such as differentiable primitive predicates and
programs~\cite{dong2018neural,gaunt17:diffprog}.
However, it is hard to design a unified differentiable module to
accurately represent general relational knowledge, which may contain complex inference
structures such as recursion~\cite{glasmachers17,garcez:19:nesy}.


Therefore, many researchers propose to break the end-to-end
learning pipeline apart, and build a hybrid model that consists of smaller
modules where each of them only accounts for one specific function~\cite{glasmachers17}.
A representative branch in this line of research is Neuro-Symbolic (NeSy)
AI~\cite{nesysurvey20,garcez:19:nesy} aiming to bridge System 1
and System 2 AI~\cite{kahneman:11:thinking,bengio:conciousness}, i.e., neural-network-based 
machine learning and symbolic-based relational inference.


However, the lack of supervision in the non-differentiable interface between
neural and symbolic systems, based on the facts extracted from raw data and their
truth values, leads to high computational complexity in
learning~\cite{qing:20:grammar,dai:abl}.
Consequently, almost all neural-symbolic models assume the existence of a very strong predefined
domain knowledge base and could not perform program induction.
This limits the expressive power of the hybrid-structured model and sacrifices
many benefits of symbolic learning (e.g., predicate invention, learning recursive
theories, and re-using learned models as background knowledge).

In this paper, we integrate neural networks with Inductive
Logic Programming (ILP)~\cite{muggleton:94:ilp} to enable first-order logic
theory induction from raw data. More specifically, we present Abductive
Meta-Interpretive Learning ($Meta_{Abd}$) which extends the Abductive Learning (ABL)
framework~\cite{dai:abl,zhou2019abductive} by combining logical induction and
abduction~\cite{flach:abductionbook} with neural networks in Meta-Interpretive Learning
(MIL)~\cite{muggleton:15:mil}. $Meta_{Abd}$ employs neural networks to extract probabilistic logic facts from
raw data, and induces an abductive logic program~\cite{KakasKT92ALP}
that can efficiently infer the truth values of the facts to train the neural
model.


To the best of our knowledge, $Meta_{Abd}$ is the first system that can
simultaneously (1) train neural models from scratch, (2) learn recursive logic theories
and (3) perform predicate invention from domains with sub-symbolic
representation. In the experiments
we compare $Meta_{Abd}$ to the compared state-of-the-art end-to-end deep
learning models and neuro-symbolic methods on two complex learning tasks. The
results show that, given the same amount of background knowledge, $Meta_{Abd}$
outperforms the compared models significantly in terms of predictive accuracy
and data efficiency, and learns human interpretable models that could be re-used
in subsequent learning tasks.

\vspace{-.5em}
\section{Related Work}
\label{sec:rel}

Solving ``System 2'' problems requires the ability of relational
and logical reasoning~\cite{kahneman:11:thinking,bengio:conciousness}.
Due to its complexity, many researchers have tried to embed
intricate background knowledge in end-to-end deep learning models. 
For example,~\cite{nalu18} propose the differentiable Neural
Arithmetic Logic Units (NALU) to model basic arithmetic functions (e.g., addition,
multiplication, etc.) in neural cells;~\cite{grover:neuralsort} encode
permutation operators with a stochastic matrix and present a differentiable
approximation to the sort operation;~\cite{Wang:19:SATnet} introduce a
differentiable SAT solver to enable gradient-based constraint solving. However,
most of these specially designed differentiable modules are \emph{ad hoc}
approximations to the original symbolic inference mechanisms.

To exploit the complex background knowledge 
expressed by formal languages directly,
Statistical Relational (StarAI) and Neural Symbolic (NeSy)
AI~\cite{nesysurvey20,garcez:19:nesy} try
to use probabilistic inference or other differentiable functions to approximate
logical inference~\cite{Cohen:20:tensorlog,dong2018neural,manhaeve:18:deepproblog,Donadello:17:ltn}.
However, they require a pre-defined symbolic knowledge base and only train
the attached neural/probabilistic models due to the highly complex interface
between the neural and symbolic modules.

One way to learn symbolic theories is to use Inductive Logic
Programming~\cite{muggleton:94:ilp}. Some
early work on combining logical abduction and induction can learn logic
theories even when input data is incomplete~\cite{flach:abductionbook}.
Recently, $\partial$ILP was proposed for learning first-order logic theories from noisy
data~\cite{evans:18:noisy}. However, ILP-based works are designed for learning in
symbolic domains. Otherwise, they need to use a fully trained neural models to
make sense of the raw inputs by extracting logical facts from the data before
program induction.

Machine apperception~\cite{evans:21:apperception} unifies Answer Set Programming
with perception by modeling it with binary neural networks. It can learn
recursive logic theories and perform concept (monadic predicate) invention.
However, both logic hypotheses and the parameters of neural networks are
represented by logical groundings, making the system very hard to optimise. For
problems involving noisy inputs like MNIST images, it still requires a fully
pre-trained neural net for pre-processing due to its high complexity in
learning.

Previous work on Abductive Learning (ABL)~\cite{dai:abl,daiLASIN} also unites
subsymbolic perception and symbolic reasoning through logical abduction, but
they need a pre-defined knowledge base to enable abduction and cannot perform
program induction. Our presented Abductive Meta-Interpretive Learning takes a
step further, which not only learns a perception model that can make sense of raw
data, but also learns logic programs and performs predicate invention to
understand the underlying relations in the task.

\begin{figure*}[t!]
  \includegraphics[width=1.0\textwidth]{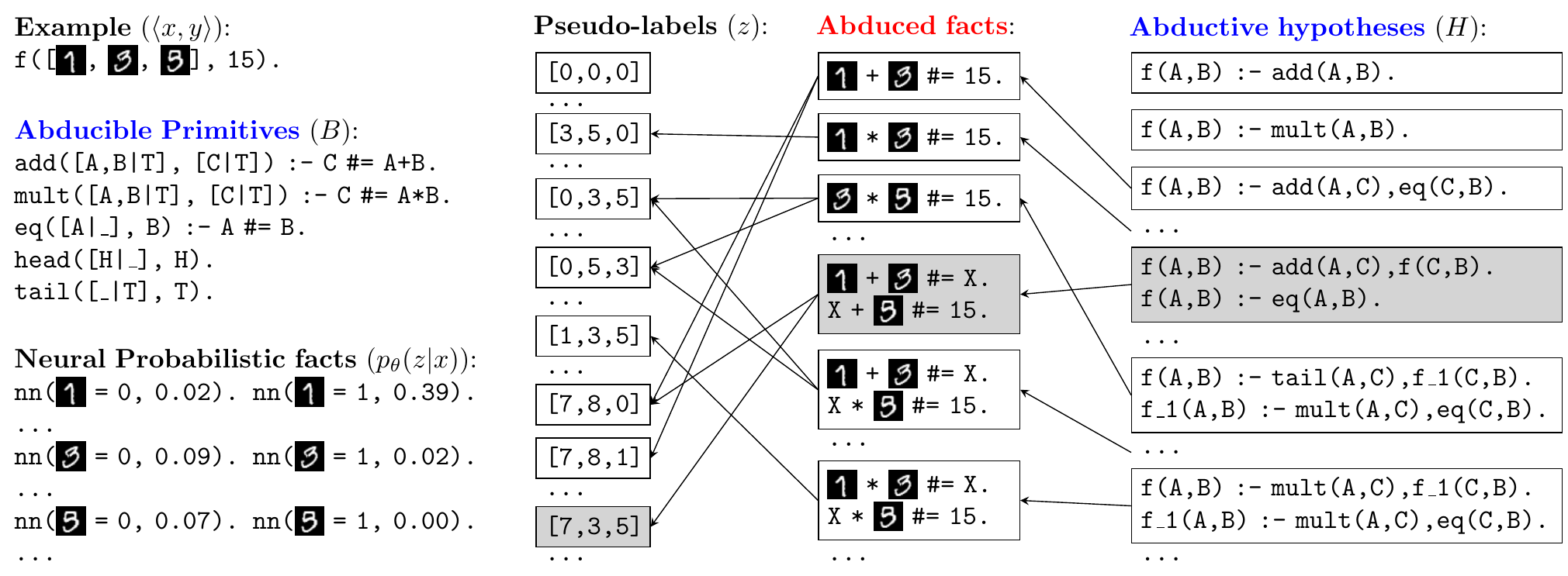}
  \caption{Example of $Meta_{Abd}$'s abduction-induction learning.
    Given training examples, background knowledge of
    abducible primitives and probabilistic facts generated by a perceptual
    neural net, $Meta_{Abd}$ learns an
    abductive logic program $H$ and abduces relational facts as constraints
    (implemented with the CLP(Z) predicate  ``\texttt{\#=}''\protect\footnotemark) over the input
    images; it then uses them to efficiently prune the search space of the most
    probable pseudo-labels $z$ (in grey blocks) for training the neural network.
    \label{fig:1}}
  \vspace{-1em}
\end{figure*}

\section{Abductive Meta-Interpretive Learning}
\label{sec:meta_abd}

\subsection{Problem Formulation}
\label{sec:setting}
A typical model bridging sub-symbolic and symbolic learning
contains two major parts: a perception model and a reasoning model~\cite{dai:abl}.
The perception model maps sub-symbolic inputs $x\in\mathcal{X}$ to some primitive
symbols $z\in\mathcal{Z}$, such as digits, objects, ground logical expressions, etc.
The reasoning model takes the interpreted $z$ as input and infers the final
output $y\in\mathcal{Y}$ according to a symbolic knowledge base $B$. Because the
primitive symbols $z$ are \emph{uncertain} and \emph{not observable} from both
training data and the knowledge base, we have named them as
\emph{pseudo-labels} of $x$.

The perception model is parameterised with $\theta$ and outputs
the conditional probability $P_\theta(z|x)=P(z|x,\theta)$; the reasoning
model $H\in\mathcal{H}$ is a set of
\emph{first-order} logical clauses such that $B\cup H\cup z\models y$, where
``$\models$'' means ``logically entails''. Our target is to learn $\theta$ and $H$
simultaneously from training data $D=\{\langle x_i, y_i\rangle\}_{i=1}^n$. For
example, if we have one example with
$x=\mathtt{[\img{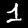},\img{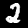},\img{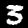}]}$ and
$y=6$, given background knowledge about adding two numbers, the hybrid model
should learn a perception model that recognises $z=\mathtt{[1,2,3]}$ and induce a
program to add each number in $z$ recursively.

Assuming that $D$ is an i.i.d. sample from the underlying distribution of
$(x,y)$, our objective can be represented as
\begin{equation}
(H^*, \theta^*)=\argmax_{H,\theta} \prod_{\langle x,y\rangle\in D}\sum_{z\in\mathcal{Z}}P(y,z|B,x,H,\theta),\label{eq:prob0}
\end{equation}
where pseudo-label $z$ is a hidden variable. Theoretically, this
problem can be solved by Expectation Maximisation (EM) algorithm. However,
the symbolic hypothesis $H$---a first-order logic theory---is difficult to
be optimised together with the parameter $\theta$, which has a continuous hypothesis
space.


We propose to solve this problem by treating $H$ like $z$ as an extra hidden variable, which gives us:
\begin{equation}
\theta^*=\argmax_\theta \prod_{\langle x,y\rangle\in D}\sum_{H\in\mathcal{H}}\sum_{z\in\mathcal{Z}}P(y,H,z|B,x,\theta).\label{eq:prob1}
\end{equation}
Now, the learning problem can be split into two EM steps:
(1) \textbf{Expectation:} obtaining the expected value of $H$ and $z$ by
sampling them in their discrete hypothesis space from ${(H,z)\sim P(H,z|B,x,y,\theta)}$;
(2) \textbf{Maximisation:} estimating $\theta$ by maximising the likelihood of
training data with numerical optimisation approaches such as gradient descent.



\paragraph{Challenges} The main challenge is to estimate the expectation of the hidden
variables $H\cup z$, i.e., we need to search for the most probable $H$ and $z$
given the $\theta$ learned in the previous iteration. This is not trivial. Even when
$B$ is sound and complete, estimating the truth-values of
hidden variable $z$ results in a search space growing exponentially with the
number of training examples, which is verified in our experiments with
DeepProblog~\cite{manhaeve:18:deepproblog} in section~\ref{sec:exp_1}.


Furthermore, the size and structure of hypothesis space $\mathcal{H}$ of
first-order logic programs makes the search problem even more complicated. For
example, given $x=\mathtt{[\img{figs/1.png},\img{figs/2.png},\img{figs/3.png}]}$ and $y=6$,
when the perception model is accurate enough to output the most probable
$z=\mathtt{[1,2,3]}$, we have at least two choices for $H$: cumulative sum or
cumulative product. When the perception model is under-trained and outputs the
most probable $z=\mathtt{[2,2,3]}$, then $H$ could be a program that only
multiplies the last two digits. Hence, $H$ and $z$ are entangled and cannot be
treated independently.

\subsection{Probabilistic Abduction-Induction Reasoning}
\label{sec:abd_ind}

Inspired by early works in abductive logic programming~\cite{flach:abductionbook}, we
propose to solve the challenges above by combining logical induction and
abduction. The induction learns an abductive logic theory $H$ based on $P_\theta(z|x)$;
the abduction made by $H$ reduces the search space of $z$.

\emph{Abductive reasoning}, or \emph{abduction} refers to the process of
selectively inferring specific grounded facts and hypotheses that give the best
explanation to observations based on background knowledge of a deductive theory.

\begin{definition}[Abducible primitive]
  An abducible primitive is a predicate that defines the explanatory grounding
  facts in abductive reasoning.
\end{definition}

\begin{definition}[Abductive hypothesis]
  An abductive hypothesis is a set of first-order logic clauses whose body
  contains literals of abductive primitives.
\end{definition}

Following is an example of using \emph{abductive hypothesis} and \emph{abducible
  primitive} in problem-solving:  

\begin{example}
  \textnormal{Observing raw inputs $x=\mathtt{[\img{figs/2.png},\img{figs/3.png},\img{figs/1.png}]}$
  and a symbolic output $y=6$, we could formulate an \emph{abductive hypothesis} $H$
  that is a recursive cumulative sum function, whose \emph{abductive primitives}
  are ``$+$'' and ``$=$''. Hence, $H$ will abduce a set of
  explanatory ground facts $\{\img{figs/2.png}+\img{figs/3.png}=\mathtt{Z},
  \mathtt{Z}+\img{figs/1.png}=6\}$. Based on these facts, we could infer that
  none of the digits in $x$ is greater than $6$. Furthermore, if the current
  perception model assigns very high probabilities to $\img{figs/2.png}=2$ and
  $\img{figs/3.png}=3$, we could easily infer that $\img{figs/1.png}=1$ even
  when the perception model has relatively low confidence about it, as this is
  the only solution that satisfies the constraint stated by the explanatory
  groundings.}
\end{example}

\footnotetext{CLP(Z) is a constraint logic programming package accessible at https://github.com/triska/clpz. 
More implementation details are in Appendix.}

An illustrative example of combining abduction and induction with probabilities
is shown in Fig.~\ref{fig:1}. Briefly speaking, instead of directly sampling
pseudo-labels $z$ and $H$ together from the huge hypothesis space, our
$Meta_{Abd}$ induces \emph{abductive hypothesis} $H$ consists of \emph{abducible
  primitives}, and then use the abduced facts to prune the search space of $z$.
Meanwhile, the perception model outputs the likelihood of pseudo-labels with
$p_\theta(z|x)$ defining a distribution over all possible values of $z$ and
helps to find the most probable $H\cup z$.

Formally, we re-write the likelihood of each $\langle x, y\rangle$ in Eq.~\ref{eq:prob1}:
\begin{eqnarray}
  P(y,H,z|B,x,\theta) = &\hspace{-5.3em} P(y,H|B,z)P_\theta(z|x)\nonumber\\
                      = &\hspace{-.9em}P(y|B,H,z)P(H|B,z)P_\theta(z|x)\nonumber\\
                      = &\hspace{-.9em}P(y|B,H,z)P_{\sigma^*}(H|B)P_\theta(z|x),\label{eq:prob3}
\end{eqnarray}
where $P_{\sigma^*}(H|B)$ is the Bayesian prior distribution on first-order
logic hypotheses, which is defined by the transitive closure of
\emph{stochastic refinements} $\sigma^*$ given the background
knowledge $B$~\cite{Muggleton:13:metabayes}, where a refinement $\sigma$ is a unit
modification (e.g., adding/removing a clause or literal) to a logic theory. The equations hold because: (1)
pseudo-label $z$ is conditioned on $x$ and $\theta$ since it is the output of the
perception model; (2) $H$ follows the prior
distribution so it only depends on $B$; (3) $y\cup H$ is independent from
$x$ given $z$ because the relations among $B$, $H$, $y$ and $z$ are determined
by pure logical inference, where:
\begin{equation}
  P(y|B,H,z) = \begin{cases}
     1, & \text{if }B\cup H\cup z\models y,\\
     0, & \text{otherwise}.
 \end{cases}\label{eq:pure}
\end{equation}

Following Bayes' rule we have
$P(H,z|B,x,y,\theta)\propto P(y,H,z|B,x,\theta)$. Now we can search for the
most probable $H\cup z$ in the expectation step according to Eq.~\ref{eq:prob3} as follows:
\begin{enumerate}
\item Induce an \emph{abductive} theory $H\sim P_{\sigma^*}(H|B)$;
\item Use $H\cup B$ and $y$ to \emph{abduce}\footnote{It can be parallelled, please
    see Appendix.} possible pseudo-labels $z$, which are
  \emph{guaranteed} to
  satisfy $H\cup B\cup z\vdash y$ and exclude the values of $z$ such that $P(y|B,H,z)=0$;
\item According to Eq.~\ref{eq:prob3} and~\ref{eq:pure}, score each sampled $H\cup z$:
  \begin{equation}
    score(H,z)=P_{\sigma^*}(H|B) P_\theta(z|x)\label{eq:score}
  \end{equation}
\item Return the $H\cup z$ with the highest score.
\end{enumerate}


\subsection{The $Meta_{Abd}$ Implementation}
\label{sec:alg}

\begin{figure}[t!]
  \centering
  \resizebox{0.47\textwidth}{!}{%
  \begin{tabular}{|l|}
    \hline
    \hspace{9.3em}\textbf{Abductive Meta-Interpreter}\\
    \hline
    \texttt{prove([], Prog, Prog, [], Prob, Prob).}\\
    \texttt{prove([Atom|As], Prog1, Prog1, Abds, Prob1, Prob2) :-}\\
    \texttt{\hspace{2.5em}deduce(Atom),}\\
    \texttt{\hspace{2.5em}prove(As, Prog1, Prog2, Abds, Prob1, Prob2).}\\
    \textbf{\texttt{prove([Atom|As], Prog1, Prog1, Abds, Prob1, Prob2) :-}}\\
    \textbf{\texttt{\hspace{2.5em}call\_abducible(Atom, Abd, Prob),}}\\
    \textbf{\texttt{\hspace{2.5em}Prob3 is Prob1 * Prob,}}\\
    \textbf{\texttt{\hspace{2.5em}get\_max\_prob(Max), Prob3 > Max,}}\\
    \textbf{\texttt{\hspace{2.5em}set\_max\_prob(Prob3),}}\\
    \textbf{\texttt{\hspace{2.5em}prove(As, Prog1, Prog1, [Abd|Abds],
    Prob3, Prob2).}}\\
    \texttt{prove([Atom|As], Prog1, Prog2, Abds, Prob1, Prob2) :-}\\
    \texttt{\hspace{2.5em}meta-rule(Name, MetaSub,(Atom :- Body), Order),}\\
    \texttt{\hspace{2.5em}Order,}\\
    \texttt{\hspace{2.5em}substitue(metasub(Name, MetaSub), Prog1, Prog3),}\\
    \texttt{\hspace{2.5em}prove(Body, Prog3, Prog4),}\\
    \texttt{\hspace{2.5em}prove(As, Prog4, Prog2, Abds, Prob1, Prob2)}\\
    \hline
  \end{tabular}
}
  \vspace{-.5em}
  \caption{Prolog code for $Meta_{Abd}$.\label{fig:meta_abd}}
  \vspace{-1.2em}
\end{figure}

We implement the above abduction-induction algorithm with Abductive
Meta-Interpretive Learning ($Meta_{Abd}$).

Meta-Interpretive Learning~\cite{muggleton:15:mil} is a form of
ILP~\cite{muggleton:94:ilp}. It learns first-order logic programs with a
second-order meta-interpreter, which consists of a definite first-order
background knowledge $B$ and meta-rules $M$. $B$ contains the primitive
predicates for constructing first-order hypotheses $H$; $M$ is second-order
clauses with existentially quantified predicate variables and universally
quantified first-order variables. In short, MIL attempts to prove the training
examples and saves the resulting programs for successful proofs.


$Meta_{Abd}$ extends the general meta-interpreter of MIL by including an
abduction procedure (bold fonts in Fig.~\ref{fig:meta_abd}) that can abduce
groundings (e.g., specific constraints on pseudo-labels $z$). As shown in
Fig.~\ref{fig:meta_abd}, it recursively proves a series of atomic goals by
deduction (\texttt{deduce/1}), abducing explanatory facts
(\texttt{call\_abducible/3}) or generating a new clause from
\texttt{meta-rule/4}.

The last argument of \texttt{call\_abducible/3}, $\mathtt{Prob}=P_\theta(z|x)$,
describes the distribution of possible worlds collected from the raw inputs.
It helps pruning the search space of the \emph{abductive hypothesis} $H$.
During the iterative refinement of $H$, $Meta_{Abd}$ greedily aborts its current
\texttt{prove/6} procedure once it has a lower probability than the best
abduction so far (the 8th line in Fig.~\ref{fig:meta_abd}).


After an abductive hypothesis $H$ has been constructed, the search for $z$
will be done by logical abduction. Finally, the score of $H\cup z$ will be calculated by
Eq.~\ref{eq:score}, where $P_\theta(z|x)$ is the output of the perception
model, which in this work is implemented with a neural network $\varphi_\theta$
that outputs:
\begin{equation*}
  P_\theta(z|x) = softmax(\varphi_\theta(x,z)).
\end{equation*}
Meanwhile, we define the prior distribution on $H$ by following~\cite{hocquette:18:activemil}:
\begin{equation*}
  P_{\sigma^*}(H|B)=\frac{6}{(\pi\cdot c(H))^2},
\end{equation*}
where $C(H)$ is the complexity of $H$, e.g., its size.

\section{Experiments}
\label{sec:exp}

This section describes the experiments of learning recursive arithmetic and
sorting algorithms from images of handwritten digits\footnote{Code \& data:
  https://github.com/AbductiveLearning/Meta\_Abd}, aiming to address
the following questions:
\begin{enumerate}
\item Can $Meta_{Abd}$ learn first-order logic programs and train perceptual
  neural networks jointly?  
\item Given the same or less amount of domain knowledge shown in
  Tab.~\ref{tab:bk}, is hybrid modelling, which directly leverages the
  background knowledge in symbolic form, better than end-to-end learning?
\end{enumerate}

\begin{table*}
  \centering
  \resizebox{1.0\textwidth}{!}{%
  \begin{tabular}{llll}
    \toprule
    \textbf{Domain Knowledge} & \textbf{End-to-end Models} & \textbf{Neuro-Symbolic Models} & $\mathbf{Meta_{Abd}}$\\
    \midrule
    Recurrence & LSTM \& RNN & Problog's list operations & Prolog's list operations\\
    \midrule
    Arithmetic functions &  \texttt{NAC}\& \texttt{NALU}~\cite{nalu18} & Full
                                                                         program
                                                                         of
                                                                         accumulative
                                                                         sum/product
                                                                 & Predicates \texttt{add},
                                                            \texttt{mult} and
                                                            \texttt{eq}\\
    \midrule
    Sequence \& Odering & Permutation matrix
                          $P_{\mathtt{sort}}$~\cite{grover:neuralsort} &
                                                                         Predicates
                                                                         ``$>$'',
                                                                         ``$=$''
                                                                 and
                                                                         ``$<$''~\cite{dong2018neural} & Prolog's
                                                                \texttt{permutation}\\
    \midrule
    Sorting & \texttt{sort} operator~\cite{grover:neuralsort} &
                                                                \texttt{swap(i,j)}
                                                                operator~\cite{dong2018neural}
                                                                 & Predicate \texttt{s}
                                                         (learned from
                                                         sub-task)\\
    \bottomrule
  \end{tabular}
}%
\vspace{-.5em}
  \caption{Domain knowledge used by the compared models.\label{tab:bk}
  }
  \vspace{-.5em}
\end{table*}

\begin{table*}
  \centering
  \resizebox{.75\textwidth}{!}{%
  \begin{tabular}{lcccccccc}
    \toprule
        & \multicolumn{4}{c}{\textbf{MNIST cumulative sum}} & \multicolumn{4}{c}{\textbf{MNIST
                                                     cumulative product}}\\

    \cmidrule(lr){2-5}
    \cmidrule(lr){6-9}
    & \textbf{Acc.} & \multicolumn{3}{c}{\textbf{MAE}} & \textbf{Acc.}
                     &\multicolumn{3}{c}{\textbf{log MAE}}\\
    \cmidrule(lr){2-2}
    \cmidrule(lr){3-5}
    \cmidrule(lr){6-6}
    \cmidrule(lr){7-9}
    Sequence Length&1&5&10&100&1&5&10&15\\
    \midrule
    LSTM &9.80\%&15.3008&44.3082&449.8304&9.80\%&11.1037&19.5594&21.6346\\
    RNN-Relu &10.32\%&12.3664&41.4368&446.9737&9.80\%&10.7635&19.8029&21.8928\\
    \midrule
    DeepProblog & \multicolumn{4}{c}{Training timeout (72 hours)} & \textbf{93.64\%} &
                                                                            \multicolumn{3}{c}{Test
                                                                                     timeout
                                                                                     (72
                                                                                       hours)}
    \\
    \midrule
    LSTM-NAC&7.02\%&6.0531&29.8749&435.4106&0.00\%&9.6164&20.9943&17.9787\\
    LSTM-NAC$_{10k}$&8.85\%&1.9013&21.4870&424.2194&10.50\%&9.3785&20.8712&17.2158\\
    LSTM-NALU&0.00\%&6.2233&32.7772&438.3457&0.00\%&9.6154&20.9961&17.9487\\
    LSTM-NALU$_{10k}$&0.00\%&6.1041&31.2402&436.8040&0.00\%&8.9741&20.9966&18.0257\\
    $Meta_{Abd}$&\textbf{95.27\%}&\textbf{0.5100}&\textbf{1.2994}&\textbf{6.5867}&\textbf{97.73\%}&\textbf{0.3340}&\textbf{0.4951}&\textbf{2.3735}\\
    \midrule
    LSTM-NAC$_{\text{1-shot CNN}}$&49.83\%&0.8737&21.1724&426.0690&0.00\%&6.0190&13.4729&17.9787\\
    LSTM-NALU$_{\text{1-shot CNN}}$&0.00\%&6.0070&30.2110&435.7494&0.00\%&9.6176&20.9298&18.1792\\
    $Meta_{Abd+\text{1-shot CNN}}$&\textbf{98.11\%}&\textbf{0.2610}&\textbf{0.6813}&\textbf{4.7090}&\textbf{97.94\%}&\textbf{0.3492}&\textbf{0.4920}&\textbf{2.4521}\\
    \bottomrule
  \end{tabular}%
}
  \vspace{-.5em}
  \caption{Accuracy on the MNIST cumulative sum/product
    tasks.~\label{tab:result1}}
  \vspace{-1.2em}
\end{table*}

\subsection{Cumulative sum and product from images}
\label{sec:exp_1}
\paragraph{Materials} We follow the settings in~\cite{nalu18}. The inputs of the
two tasks are sequences of randomly chosen MNIST digits; the numerical outputs are
the sum and product of the digits, respectively. The lengths of training
sequences are 2--5. To verify if the learned models can extrapolate to longer
inputs, the length of test examples ranges from 5 to 100.
For cumulative product, when the randomly generated sequence is
long enough, it will be very likely to contain a $0$ and makes the final outputs
equal to $0$. So the extrapolation examples has maximum length 15 and only
contain digits from 1 to 9. The dataset contains 3000 and 1000 examples for
training and validation, respectively; the test data of each length has 10,000
examples.


\paragraph{Methods}
We compare $Meta_{Abd}$ with following state-of-the-art baselines: End-to-end
models include RNN, LSTM and LSTMs attached to Neural Accumulators(NAC) and
Neural Arithmetic Logic Units (NALU)~\cite{nalu18}; NeSy system
DeepProblog~\cite{manhaeve:18:deepproblog}\footnote{We use NAC/NALU at
  https://github.com/kevinzakka/NALU-pytorch; DeepProblog at  
https://bitbucket.org/problog/deepproblog}.

A convnet processes the input images to the recurrent networks and Problog
programs, as~\cite{nalu18} and~\cite{manhaeve:18:deepproblog} described;
it also serves as the perception model of $Meta_{Abd}$ to output
the probabilistic facts. As shown in Tab.~\ref{tab:bk}, NAC, NALU and
$Meta_{Abd}$ are aware of the same amount of background knowledge for learning
both perceptual convnet and recursive arithmetic algorithms jointly, while
DeepProblog is provided with the ground-truth program and only trains the
perceptual convnet. Like NAC and NALU, $Meta_{Abd}$ uses the same background
knowledge for both sum and product tasks.

Each experiment is carried out five
times, and the average of the results are reported. The performance is measured
by classification accuracy (Acc.) on length-one inputs, mean average error (MAE)
in sum tasks, and mean average error on logarithm (log MAE) of the outputs in
product tasks whose error grows exponentially with sequence length. 

\begin{wrapfigure}{R}{0.24\textwidth}
  \centering
  \begin{subfigure}[b]{0.24\textwidth}
    \centering
    \includegraphics[width=1.0\textwidth]{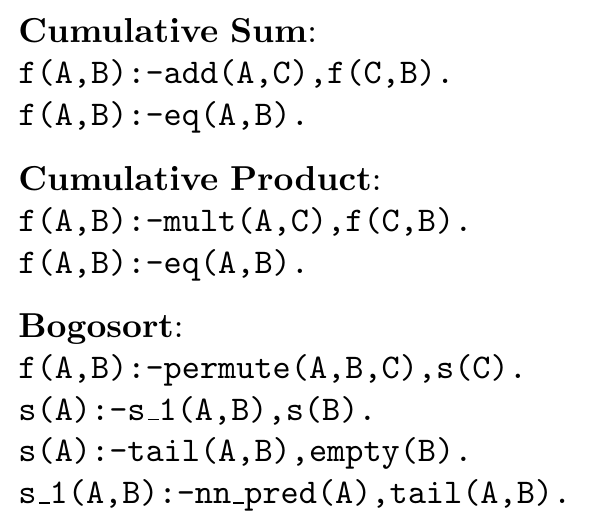}%
    \vspace{-.5em}
    \caption{Learned programs\label{fig:prog}}%
  \end{subfigure}
  \begin{subfigure}[b]{0.24\textwidth}
    \centering
    \includegraphics[width=\textwidth]{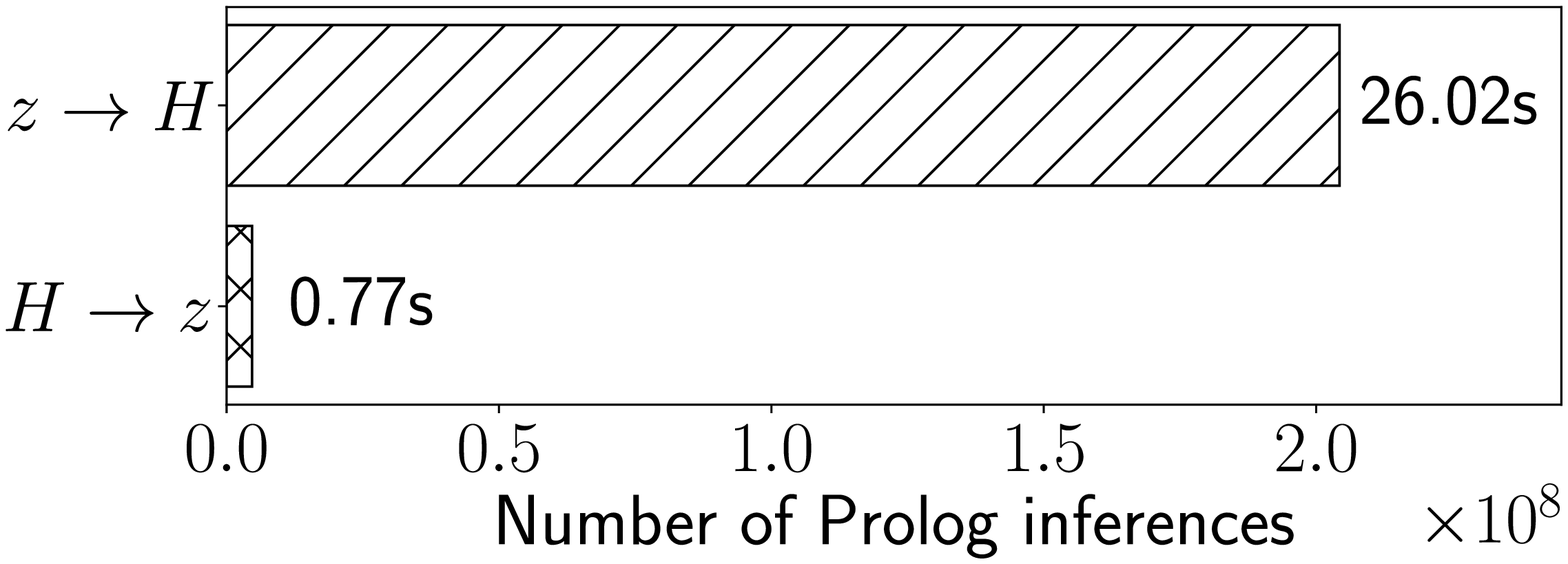}
    \caption{Time costs of sampling $z$ or $H$\label{fig:time}}
  \end{subfigure}
  \caption{Learned programs and the time efficiency of $Meta_{Abd}$.}
\end{wrapfigure}

\paragraph{Results} Our experimental results are shown in
Tab.~\ref{tab:result1}; the learned first-order logic theories are shown in
Fig.~\ref{fig:prog}. The end-to-end models that do not exploit any background
knowledge (LSTM and RNN) perform worst. NALU and NAC is slightly better because
they include neural cells with arithmetic modules, but the end-to-end learning
pipeline based on embeddings results in low sample-efficiency. DeepProblog does
not finish the training on the cumulative sum task and the test on cumulative
product task within 72 hours because the recursive programs result in a
huge groundings space for its maximum a posteriori (MAP) estimation.

\begin{table*}[ht!]
  \centering
  \resizebox{.6\textwidth}{!}{%
    \begin{tabular}{lcccc}
      \toprule
      \textbf{Sequence Length} & \textbf{3} & \textbf{5} & \textbf{7}\\
      \midrule
      Neural Logical Machine (NLM) & 17.97\% (34.38\%) & 1.03\% (20.27\%) & 0.01\% (14.90\%)\\
      \midrule
      Deterministic NeuralSort &95.49\% (96.82\%)&88.26\% (94.32\%)&80.51\% (92.38\%)\\
      \midrule
      Stochastic NeuralSort &95.37\% (96.74\%)&87.46\% (94.03\%)&78.50\% (91.85\%)\\
      \midrule
      $Meta_{Abd}$&\textbf{96.33\% (97.22\%)}&\textbf{91.75\% (95.24\%)}& \textbf{87.42\% (93.58\%)}\\
      \bottomrule
    \end{tabular}
  }%
  \vspace{-.5em}
  \caption{Accuracy of MNIST sort. First value is the rate of correct permutations; 
    second value is the rate of correct individual element
    ranks.\label{tab:result2}}
  \vspace{-1.2em}
\end{table*}

\setlength{\intextsep}{0pt}%
\setlength{\columnsep}{5pt}%
The EM-based learning of $Meta_{Abd}$ may be trapped in local optima, which
happens more frequently in cumulative sum than produce since its distribution
$P(H,z|B,x,y,\theta)$ is much denser.
Therefore, we also carry out experiments with one-shot pre-trained convnets,
which are trained by randomly sampling one example in each
class from MNIST data. Although the pre-trained convnet is weak
at start (Acc. 20\%$\sim$35\%), it provides a good
initialisation and significantly improves the learning
performance.

Fig.~\ref{fig:time} compares the time efficiency between ILP's induction and
$Meta_{Abd}$'s abduction-induction in one EM iteration of learning
cumulative sum. ``$z\rightarrow H$'' means first sampling $z$ and then inducing $H$ with
ILP; ``$H\rightarrow z$'' means first sampling an abductive hypothesis $H$ and then using
$H$ to abduce $z$. The x-axis denotes the average number of Prolog inferences,
the number at the end of each bar is the average inference time in seconds. Evidently, the
abduction leads to a substantial improvement in the number of Prolog inferences
and significantly the complexity of searching pseudo-labels.

\setlength{\intextsep}{10pt}%
\setlength{\columnsep}{10pt}%

\vspace{-.8em}
\subsection{Bogosort from images}
\label{sec:exp_2}

\paragraph{Materials} We follow the settings in~\cite{grover:neuralsort}. The
input of this task is a sequence of randomly chosen MNIST images of distinct numbers; the output
is the correct ranking (from large to small) of the digits. For example, when
$x=\mathtt{[\img{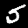},\img{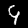},\img{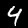},\img{figs/3.png},\img{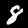}]}$
then the output should be $y=\mathtt{[3,1,4,5,2]}$ because the ground-truth
labels $z^*=\mathtt{[5,9,4,3,8]}$. The training
dataset contains 3000 training/test and 1000 validation examples. The training
examples are sequences of length 5, and we test the learned models on image
sequences with lengths 3, 5 and 7. 

\paragraph{Methods} We compare $Meta_{Abd}$ to an end-to-end model
NeuralSort~\cite{grover:neuralsort} and a state-of-the-art NeSy approach Neural
Logical Machines (NLM)~\cite{dong2018neural}\footnote{We use NeuralSort
  at https://github.com/ermongroup/neuralsort; NLM at
  https://github.com/google/neural-logic-machines.}. All experiments
are repeated five times.

NeuralSort can be regarded as a differentiable approximation to bogosort (permutation
sort). Given an input list of scalars, it generates a stochastic permutation
matrix by applying the pre-defined deterministic or stochastic \texttt{sort}
operator on the inputs. NLM can learn sorting through reinforcement learning in
a domain whose states are described by vectors of relational features
(groundings of dyadic predicates``$>$'', ``$==$'', ``$<$'') and action
``\texttt{swap}''. However, the original NLM only takes symbolic
inputs\footnote{Please see https://github.com/google/neural-logic-machines\\/blob/master/scripts/graph/learn\_policy.py},
which provides a noisy-free relational features vector. In our experiments, we attach
NLM with the same convnet as other methods to process images. We also compared to
DeepProblog with the ground-truth program of sorting in this task, but it does
not terminate when the neural predicate
``\texttt{swap\_net}''\footnote{Please see https://bitbucket.org/problog/deepproblog/src/master\\/examples/NIPS/Forth/Sort/quicksort.pl}
is implemented to take noisy image inputs by the aforementioned convnet.
Therefore, we do not display its performance in this task.

For $Meta_{Abd}$, it is easy to include stronger background knowledge 
for learning more efficient sorting algorithms~\cite{cropper:19:efficientMIL}. 
But in order to make a fair comparison to NeuralSort, we adapt the same background
knowledge to logic program and let $Meta_{Abd}$ learn bogosort. 
The knowledge of permutation in $Meta_{Abd}$ is implemented with
Prolog's built-in predicate \texttt{permutation}. Meanwhile, instead of
providing the information about
sorting as prior knowledge like the NeuralSort, we try to \emph{learn} the concept of ``sorted''
(represented by a monadic predicate \texttt{s}) from data as a sub-task, whose training set 
is the subset of the sorted examples within the
training dataset ($<20$ examples). The two tasks are trained
sequentially as a curriculum. $Meta_{Abd}$ learns the sub-task in the first five epochs and then
re-uses the learned models to learn bogosort.

$Meta_{Abd}$ uses an MLP attached to the same untrained convnet as other models
to produce dyadic probabilistic facts
\texttt{nn\_pred([\img{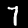},\img{figs/9.png}|\_])},
which learns if the first two items in the image sequence satisfy a dyadic
relation. Unlike NLM, the background knowledge of $Meta_{Abd}$ is agnostic to
ordering, i.e., the dyadic \texttt{nn\_pred} is not provided with supervision on whether
it should learn ``greater than'' or ``less than'', so \texttt{nn\_pred} only learns 
an unknown dyadic partial order among MNIST images. As we can see, the background knowledge
used by $Meta_{Abd}$ is \emph{much weaker} than the others.

\paragraph{Results} Tab.~\ref{tab:result2} shows the average accuracy of the compared 
methods in the sorting tasks; Fig.~\ref{fig:prog} shows the learned programs by
$Meta_{Abd}$. The performance is measured by the average proportion of correct
permutations and individual permutations following~\cite{grover:neuralsort}.
Although using weaker background knowledge, $Meta_{Abd}$ has a significantly
better performance than NeuralSort. Due to the high sample-complexity of
reinforcement learning, NLM failed to learn any valid perceptual model and
sorting algorithm (success trajectory rate 0.0\% during training).

The learned program of \texttt{s} and the dyadic neural net \texttt{nn\_pred} are
both successfully re-used in the sorting task, where the learned program of
\texttt{s} is consulted as interpreted background
knowledge~\cite{cropper:20:metaho}, and the neural network that generates
probabilistic facts of \texttt{nn\_pred} is directly re-used and continuously
trained during the learning of sorting. This experiment also demonstrates
$Meta_{Abd}$'s ability of learning recursive logic programs and predicate
invention (the invented predicate \texttt{s\_1} in Fig.~\ref{fig:prog}).



\vspace{-.8em}
\section{Conclusion}
\label{sec:conc}

In this paper, we present the Abductive Meta-Interpretive Learning ($Meta_{Abd}$) approach
that can simultaneously train neural networks and learn recursive first-order
logic theories with predicate invention. By combining ILP with neural networks,
$Meta_{Abd}$ can learn human-interpretable logic programs directly from
raw-data, and the learned neural models and logic theories can be directly re-used
in subsequent learning tasks. $Meta_{Abd}$ adopts a general framework for
combining perception with logical induction and abduction.
The perception model extracts probabilistic facts from sub-symbolic data; the
logical induction searches for first-order abductive theories in a relatively
small hypothesis space; the logical abduction uses the abductive theory to prune
the vast search space of the truth values of the probabilistic facts. The three
parts are optimised together in a probabilistic model.

In future work, we would like to apply $Meta_{Abd}$
in real tasks such as computational science discovery, which is a typical
abductive process that involve both symbolic domain knowledge and
continuous/noisy raw data. Since $Meta_{Abd}$ uses pure logical
inference for reasoning, it is possible to leverage more advanced symbolic
inference/optimisation techniques like Satisfiability Modulo Theories
(SMT)~\cite{barrett:18:smt} and Answer Set Programming
(ASP)~\cite{lifschitz:19:asp} to reason more efficiently.

\vspace{-.8em}
\section*{Acknowledgements}
The first author acknowledges support from the UK's EPSRC Robot Synthetic
Biologist, grant EP/R034915/1, for financial support. The second author
acknowledges support from the UK’s EPSRC Human-Like Computing Network, grant
EP/R022291/1, for which he acts as director. The authors thank C\'eline Hocquette,
Stassa Patsantzis and Ai Lun for their careful proofreading and helpful
comments.

\small
\bibliographystyle{named}
\bibliography{meta_abd}

\begin{thebibliography}{}

\bibitem[\protect\citeauthoryear{Barrett and Tinelli}{2018}]{barrett:18:smt}
Clark~W. Barrett and Cesare Tinelli.
\newblock Satisfiability modulo theories.
\newblock In {\em Handbook of Model Checking}, pages 305--343. Springer, 2018.

\bibitem[\protect\citeauthoryear{Bengio}{2017}]{bengio:conciousness}
Yoshua Bengio.
\newblock The consciousness prior.
\newblock {\em CoRR}, abs/1709.08568, 2017.

\bibitem[\protect\citeauthoryear{Cohen \bgroup \em et al.\egroup
  }{2020}]{Cohen:20:tensorlog}
William~W. Cohen, Fan Yang, and Kathryn Mazaitis.
\newblock Tensorlog: {A} probabilistic database implemented using deep-learning
  infrastructure.
\newblock {\em Journal of Artificial Intelligence Research}, 67:285--325, 2020.

\bibitem[\protect\citeauthoryear{Cropper and
  Muggleton}{2019}]{cropper:19:efficientMIL}
Andrew Cropper and Stephen~H. Muggleton.
\newblock Learning efficient logic programs.
\newblock {\em Maching Learning}, 108(7):1063--1083, 2019.

\bibitem[\protect\citeauthoryear{Cropper \bgroup \em et al.\egroup
  }{2020}]{cropper:20:metaho}
Andrew Cropper, Rolf Morel, and Stephen Muggleton.
\newblock Learning higher-order logic programs.
\newblock {\em Maching Learning}, 109(7):1289--1322, 2020.

\bibitem[\protect\citeauthoryear{Dai and Zhou}{2017}]{daiLASIN}
W.{-}Z. Dai and Z.{-}H. Zhou.
\newblock Combining logical abduction and statistical induction: Discovering
  written primitives with human knowledge.
\newblock In {\em Proceedings of the 31st {AAAI} Conference on Artificial
  Intelligence}, pages 4392--4398, San Francisco, CA, 2017.

\bibitem[\protect\citeauthoryear{Dai \bgroup \em et al.\egroup
  }{2019}]{dai:abl}
Wang{-}Zhou Dai, Qiu{-}Ling Xu, Yang Yu, and Zhi{-}Hua Zhou.
\newblock Bridging machine learning and logical reasoning by abductive
  learning.
\newblock In {\em Advances in Neural Information Processing Systems 32}, pages
  2811--2822. Curran Associates, Inc., 2019.

\bibitem[\protect\citeauthoryear{De~Raedt \bgroup \em et al.\egroup
  }{2020}]{nesysurvey20}
Luc De~Raedt, Sebastijan Dumančić, Robin Manhaeve, and Giuseppe Marra.
\newblock From statistical relational to neuro-symbolic artificial
  intelligence.
\newblock In Christian Bessiere, editor, {\em Proceedings of the 29th
  International Joint Conference on Artificial Intelligence}, pages 4943--4950.
  {IJCAI}, 7 2020.

\bibitem[\protect\citeauthoryear{Dhingra \bgroup \em et al.\egroup
  }{2020}]{dhingra:differentiable}
Bhuwan Dhingra, Manzil Zaheer, Vidhisha Balachandran, Graham Neubig, Ruslan
  Salakhutdinov, and William~W. Cohen.
\newblock Differentiable reasoning over a virtual knowledge base.
\newblock In {\em International Conference on Learning Representations}, Addis
  Ababa, Ethiopia, 2020. OpenReview.

\bibitem[\protect\citeauthoryear{Donadello \bgroup \em et al.\egroup
  }{2017}]{Donadello:17:ltn}
Ivan Donadello, Luciano Serafini, and Artur~S. d'Avila Garcez.
\newblock Logic tensor networks for semantic image interpretation.
\newblock In {\em Proceedings of the 26th International Joint Conference on
  Artificial Intelligence}, pages 1596--1602, Melbourne, Australia, 2017.
  {IJCAI}.

\bibitem[\protect\citeauthoryear{Dong \bgroup \em et al.\egroup
  }{2019}]{dong2018neural}
Honghua Dong, Jiayuan Mao, Tian Lin, Chong Wang, Lihong Li, and Denny Zhou.
\newblock Neural logic machines.
\newblock In {\em International Conference on Learning Representations}, New
  Orleans, LA, 2019. OpenReview.

\bibitem[\protect\citeauthoryear{Evans and Grefenstette}{2018}]{evans:18:noisy}
Richard Evans and Edward Grefenstette.
\newblock Learning explanatory rules from noisy data.
\newblock {\em Journal of Artificial Intelligence Research}, 61:1--64, 2018.

\bibitem[\protect\citeauthoryear{Evans \bgroup \em et al.\egroup
  }{2021}]{evans:21:apperception}
Richard Evans, Matko Bo\v{s}njak, Lars Buesing, Kevin Ellis, David Pfau,
  Pushmeet Kohli, and Marek~J. Sergot.
\newblock Making sense of raw input.
\newblock {\em Artificial Intelligence}, 299:103521, 2021.

\bibitem[\protect\citeauthoryear{Flach \bgroup \em et al.\egroup
  }{2000}]{flach:abductionbook}
Peter~A. Flach, Antonis~C. Kakas, and Antonis~M. Hadjiantonis, editors.
\newblock {\em Abduction and Induction: Essays on Their Relation and
  Integration}.
\newblock Applied Logic Series. Springer Netherlands, 2000.

\bibitem[\protect\citeauthoryear{Garcez \bgroup \em et al.\egroup
  }{2019}]{garcez:19:nesy}
Artur S.~d'Avila Garcez, Marco Gori, Lu{\'{\i}}s~C. Lamb, Luciano Serafini,
  Michael Spranger, and Son~N. Tran.
\newblock Neural-symbolic computing: An effective methodology for principled
  integration of machine learning and reasoning.
\newblock {\em IfCoLog Journal of Logics and their Applications},
  6(4):611--632, 2019.

\bibitem[\protect\citeauthoryear{Gaunt \bgroup \em et al.\egroup
  }{2017}]{gaunt17:diffprog}
Alexander~L. Gaunt, Marc Brockschmidt, Nate Kushman, and Daniel Tarlow.
\newblock Differentiable programs with neural libraries.
\newblock In {\em Proceedings of the 34th International Conference on Machine
  Learning}, volume~70, pages 1213--1222, Sydney, Australia, 2017. PMLR.

\bibitem[\protect\citeauthoryear{Glasmachers}{2017}]{glasmachers17}
Tobias Glasmachers.
\newblock Limits of end-to-end learning.
\newblock In {\em Proceedings of The 9th Asian Conference on Machine Learning},
  volume~77, pages 17--32, Seoul, Korea, 2017. {PMLR}.

\bibitem[\protect\citeauthoryear{Grover \bgroup \em et al.\egroup
  }{2019}]{grover:neuralsort}
Aditya Grover, Eric Wang, Aaron Zweig, and Stefano Ermon.
\newblock Stochastic optimization of sorting networks via continuous
  relaxations.
\newblock In {\em International Conference on Learning Representations}, New
  Orleans, LA, 2019. Openreview.

\bibitem[\protect\citeauthoryear{Hocquette and
  Muggleton}{2018}]{hocquette:18:activemil}
C{\'{e}}line Hocquette and Stephen~H. Muggleton.
\newblock How much can experimental cost be reduced in active learning of agent
  strategies?
\newblock In {\em Proceedings of the 28th International Conference on Inductive
  Logic Programming}, volume 11105, pages 38--53, Ferrara, Italy, 2018.
  Springer.

\bibitem[\protect\citeauthoryear{Kahneman}{2011}]{kahneman:11:thinking}
Daniel Kahneman.
\newblock {\em Thinking, fast and slow}.
\newblock Farrar, Straus and Giroux, New York, 2011.

\bibitem[\protect\citeauthoryear{Kakas \bgroup \em et al.\egroup
  }{1992}]{KakasKT92ALP}
Antonis~C. Kakas, Robert~A. Kowalski, and Francesca Toni.
\newblock Abductive logic programming.
\newblock {\em Journal of Logic Computation}, 2(6):719--770, 1992.

\bibitem[\protect\citeauthoryear{Li \bgroup \em et al.\egroup
  }{2020}]{qing:20:grammar}
Qing Li, Siyuan Huang, Yining Hong, Yixin Chen, Ying~Nian Wu, and Song{-}Chun
  Zhu.
\newblock Closed loop neural-symbolic learning via integrating neural
  perception, grammar parsing, and symbolic reasoning.
\newblock In {\em Proceedings of the 37th International Conference on Machine
  Learning}, volume 119, pages 5884--5894, Online, 2020. {PMLR}.

\bibitem[\protect\citeauthoryear{Lifschitz}{2019}]{lifschitz:19:asp}
Vladimir Lifschitz.
\newblock {\em Answer Set Programming}.
\newblock Springer, 2019.

\bibitem[\protect\citeauthoryear{Manhaeve \bgroup \em et al.\egroup
  }{2018}]{manhaeve:18:deepproblog}
Robin Manhaeve, Sebastijan Dumancic, Angelika Kimmig, Thomas Demeester, and Luc
  De~Raedt.
\newblock Deepproblog: Neural probabilistic logic programming.
\newblock In {\em Advances in Neural Information Processing Systems 31}, pages
  3753--3763, Montr{\'{e}}al, Canada, 2018. Curran Associates, Inc.

\bibitem[\protect\citeauthoryear{Muggleton and de
  Raedt}{1994}]{muggleton:94:ilp}
Stephen~H. Muggleton and Luc de~Raedt.
\newblock Inductive logic programming: Theory and methods.
\newblock {\em The Journal of Logic Programming}, 19-20:629 -- 679, 1994.

\bibitem[\protect\citeauthoryear{Muggleton \bgroup \em et al.\egroup
  }{2013}]{Muggleton:13:metabayes}
Stephen~H. Muggleton, Dianhuan Lin, Jianzhong Chen, and Alireza
  Tamaddoni{-}Nezhad.
\newblock {MetaBayes}: {Bayesian} meta-interpretative learning using
  higher-order stochastic refinement.
\newblock In {\em Proceedings of the 23rd International Conference on Inductive
  Logic Programming}, volume 8812, pages 1--17, Rio de Janeiro, Brazil, 2013.
  Springer.

\bibitem[\protect\citeauthoryear{Muggleton \bgroup \em et al.\egroup
  }{2015}]{muggleton:15:mil}
Stephen~H. Muggleton, Dianhuan Lin, and Alireza Tamaddoni{-}Nezhad.
\newblock Meta-interpretive learning of higher-order dyadic datalog: predicate
  invention revisited.
\newblock {\em Machine Learning}, 100(1):49--73, 2015.

\bibitem[\protect\citeauthoryear{Trask \bgroup \em et al.\egroup
  }{2018}]{nalu18}
Andrew Trask, Felix Hill, Scott~E Reed, Jack Rae, Chris Dyer, and Phil Blunsom.
\newblock Neural arithmetic logic units.
\newblock In {\em Advances in Neural Information Processing Systems 31}, pages
  8035--8044. Curran Associates, Inc., 2018.

\bibitem[\protect\citeauthoryear{Wang \bgroup \em et al.\egroup
  }{2019}]{Wang:19:SATnet}
Po{-}Wei Wang, Priya~L. Donti, Bryan Wilder, and J.~Zico Kolter.
\newblock {SATNet}: Bridging deep learning and logical reasoning using a
  differentiable satisfiability solver.
\newblock In {\em Proceedings of the 36th International Conference on Machine
  Learning}, pages 6545--6554, Long Beach, CA, 2019. {PMLR}.

\bibitem[\protect\citeauthoryear{Zhou \bgroup \em et al.\egroup
  }{2018}]{zhou18:gnnsurvey}
Jie Zhou, Ganqu Cui, Zhengyan Zhang, Cheng Yang, Zhiyuan Liu, and Maosong Sun.
\newblock Graph neural networks: {A} review of methods and applications.
\newblock {\em CoRR}, abs/1812.08434, 2018.

\bibitem[\protect\citeauthoryear{Zhou}{2019}]{zhou2019abductive}
Zhi-Hua Zhou.
\newblock Abductive learning: towards bridging machine learning and logical
  reasoning.
\newblock {\em Science China Information Sciences}, 62(7), 2019.

\end{thebibliography}

\clearpage
\appendix
\large
\onecolumn
\section{Appendix}
We introduce more implementation details and experimental results in the following sub-sections.
\subsection{Parallel Abduction}
As described in section~\ref{sec:abd_ind}, $Meta_{Abd}$ tries to estimate the most probable $z$ by abduction
following Eq.~\ref{eq:prob3}. Given training data $D=\{\langle x_i,
y_i\rangle\}_{i=1}^n$, let $\bm{x}=(x_1,\ldots,x_n)$, $\bm{y}=(y_1,\ldots,y_n)$ and
$\bm{z}=(z_1,\ldots,z_n)$, for $H\cup\bm{z}$ the posterior
$P(H,\bm{z}|B,\bm{x},\bm{y},\theta) \propto P(H,\bm{y},\bm{z}|B,\bm{x},\theta)$, which can
be further re-written as:
\begin{eqnarray}
P(\bm{y}|B,H,\bm{z})P_{\sigma^*}(H|B)P_\theta(\bm{z}|\bm{x}) & \nonumber\\
&\hspace{-10em}= P_{\sigma^*}(H|B)\prod_{i=1}^n P(y_i|B,H,z_i)P_\theta(z_i|x_i),\label{eq:para}
\end{eqnarray}
where the last equation holds because the examples are drawn i.i.d. from the underlying distribution. 

Therefore, the logical abduction in the expectation step of $Meta_{Abd}$ can be parallelised naturally:
\begin{enumerate}
    \item Sample an abductive hypothesis $H$ from the prior distribution $H\sim P_{\sigma^*}(H|B)$;
    \item Parallelly abduce $z_i$ from $H$ and $\langle x_i, y_i\rangle$, and then calculate
      their scores by Eq.~\ref{eq:score};
    \item Aggregate the results by Eq.~\ref{eq:para};
    \item Get the best $H\cup\bm{z}$ and continue the maximisation step to optimise $\theta$.
\end{enumerate}
We applied this strategy in our implementation of $Meta_{Abd}$ and have achieved better efficiency on
multi-threaded CPUs.

\subsection{MNIST Cumulative Sum/Product}
\label{sec:impl_sp}

The background knowledge used in the MNIST cumulative sum/product experiments is shown in Fig.~\ref{fig:BK1}. We
demonstrate how it works by the following example.

\begin{figure}[h]
    \centering
    \small
    \resizebox{0.48\textwidth}{!}{
    \begin{tabular}{|l|}
        \hline
        \texttt{\it \% Non-abducible primitives of list operations.}\\
        \texttt{head([H|\_],H).}\\
        \texttt{tail([\_|T],T).}\\
        \texttt{empty([]).}\\
        \\
        \texttt{\it \% {\color{red}Abducible} primitives for generating CLP constraints.}\\
        \texttt{abduce\_{\color{red}add}([X,Y|T],[N|T],{\color{blue}Abduced},1.0):-}\\
        \texttt{\hspace{2.5em}(not(ground(N)) ->}\\
        \texttt{\hspace{5.0em}metagol:new\_var(N); number(N)),}\\
        \texttt{\hspace{2.5em}atomics\_to\_string([X,'{\color{red}+}',Y,'{\color{red}\#=}',N], {\color{blue}Abduced}).}\\
        \texttt{abduce\_{\color{red}mult}([X,Y|T],[N|T],{\color{blue}Abduced},1.0):-}\\
        \texttt{\hspace{2.5em}(not(ground(N)) ->}\\
        \texttt{\hspace{5.0em}metagol:new\_var(N); number(N)),}\\
        \texttt{\hspace{2.5em}atomics\_to\_string([X,'{\color{red}*}',Y,'{\color{red}\#=}',N], {\color{blue}Abduced}).}\\
        \texttt{abduce\_{\color{red}eq}([X|T],[N|T],{\color{blue}Abduced},1.0):-}\\
        \texttt{\hspace{2.5em}(not(ground(N)) ->}\\
        \texttt{\hspace{5.0em}metagol:new\_var(N); number(N)),}\\
        \texttt{\hspace{2.5em}atomics\_to\_string([X,'{\color{red}\#=}',N], {\color{blue}Abduced}).}\\
        \hline
    \end{tabular}%
    }
    \caption{Background knowledge used in the MNIST cumulative sum/product tasks.\label{fig:BK1}}
\end{figure}

\paragraph{Example (Constraint abduction)}
Given a training example \texttt{f([\img{figs/1.png},\img{figs/2.png},\img{figs/3.png},\img{figs/9.png}],15)},
$Meta_{Abd}$ will try to learn a program of the dyadic predicate \texttt{f} to satisfy (i.e., logically prove) the example.
The program to be learned is the \emph{abductive} hypothesis $H$. The learning process is similar to
generic Meta-Interpretive Learning~\cite{muggleton:15:mil} except that it abduces some ground expressions
(the \texttt{\color{blue}Abduced} atom in Fig.~\ref{fig:BK1}) according to the definition of the 
abducible primitives. In the MNIST sum/product tasks,
the \texttt{\color{blue}Abduced} atoms are strings like ``\texttt{X+\img{figs/2.png}\#=3}'', which
is a CLP(Z)\footnote{https://github.com/triska/clpz} constraint. According to the definition in Fig.~\ref{fig:BK1},
when the Prolog variable is not grounded (i.e., constant), the abducible variable will create a new variable
to represent \texttt{N}; if the Prolog variable is grounded to a number, which means it is the 
final output in our example, then there is no need to generate a new variable to represent it. 
Assume that the currently sampled $H$ is the cumulative
sum program in Fig.~\ref{fig:prog}, then for the example \texttt{f([\img{figs/1.png},\img{figs/2.png},\img{figs/3.png},\img{figs/9.png}],15)}
$Meta_{Abd}$ can abduce four CLP(Z) constraints: ``\texttt{\img{figs/1.png}+\img{figs/2.png}\#=N1}'',
``\texttt{N1+\img{figs/3.png}\#=N2}'', ``\texttt{N2+\img{figs/9.png}\#=N3}'' and ``\texttt{N3\#=15}''. Note that
the scores of the abducibles in Fig.~\ref{fig:BK1} are all $1.0$, which means that these constraints are
\emph{hard constraints} that have to be satisfied.

After abducing the constraints, $Meta_{Abd}$ will call the CLP(Z) to solve them, giving
a small set of pseudo-labels $z$ that satisfy those constraints. Then, $Meta_{Abd}$ will try to calculate the scores
of the abduced $H\cup z$ according to Eq.~\ref{eq:score}. $P_{\sigma^*}(H|B)$ is directly given by
$H$'s complexity, i.e., the size of the program; $P_\theta(z|x)$ is given by the probabilistic facts by the
perception neural network, which are shown in Fig.~\ref{fig:facts1}. The predicate ``\texttt{nn(Img,Label,Prob)}''
means the probability of \texttt{Img} being an instance of \texttt{Label} is \texttt{Prob}. To get the 
probability of all pseudo-labels of an image sequence, $Meta_{Abd}$ simply multiplies the probabilities
of each image:
\begin{equation}
    p_\theta(z|x)=\prod_{j}p_\theta(z_j|x_j),\nonumber
\end{equation}
where $x_j$ is the $j$-th image in $x$ (first argument of predicate \texttt{nn}), $z_j$ is the abduced pseudo-label of $x_j$
(second argument of \texttt{nn}), and the probability is the third argument of \texttt{nn}.
\begin{figure*}[!t]
    \centering
    \begin{subfigure}[b]{0.45\textwidth}
    \includegraphics[width=\textwidth]{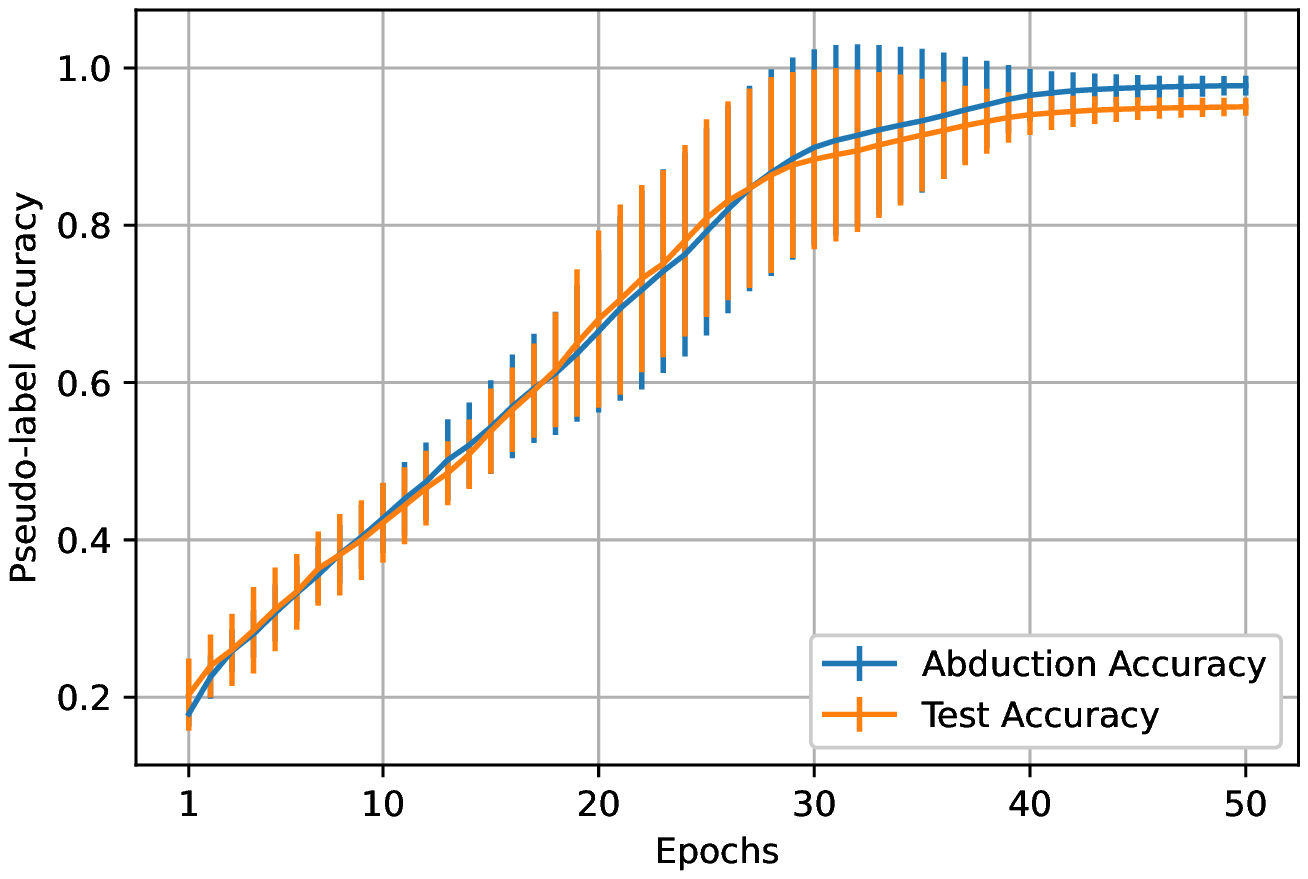}
    \caption{MNIST sum}
    \end{subfigure}
    ~
    \begin{subfigure}[b]{0.45\textwidth}
    \includegraphics[width=\textwidth]{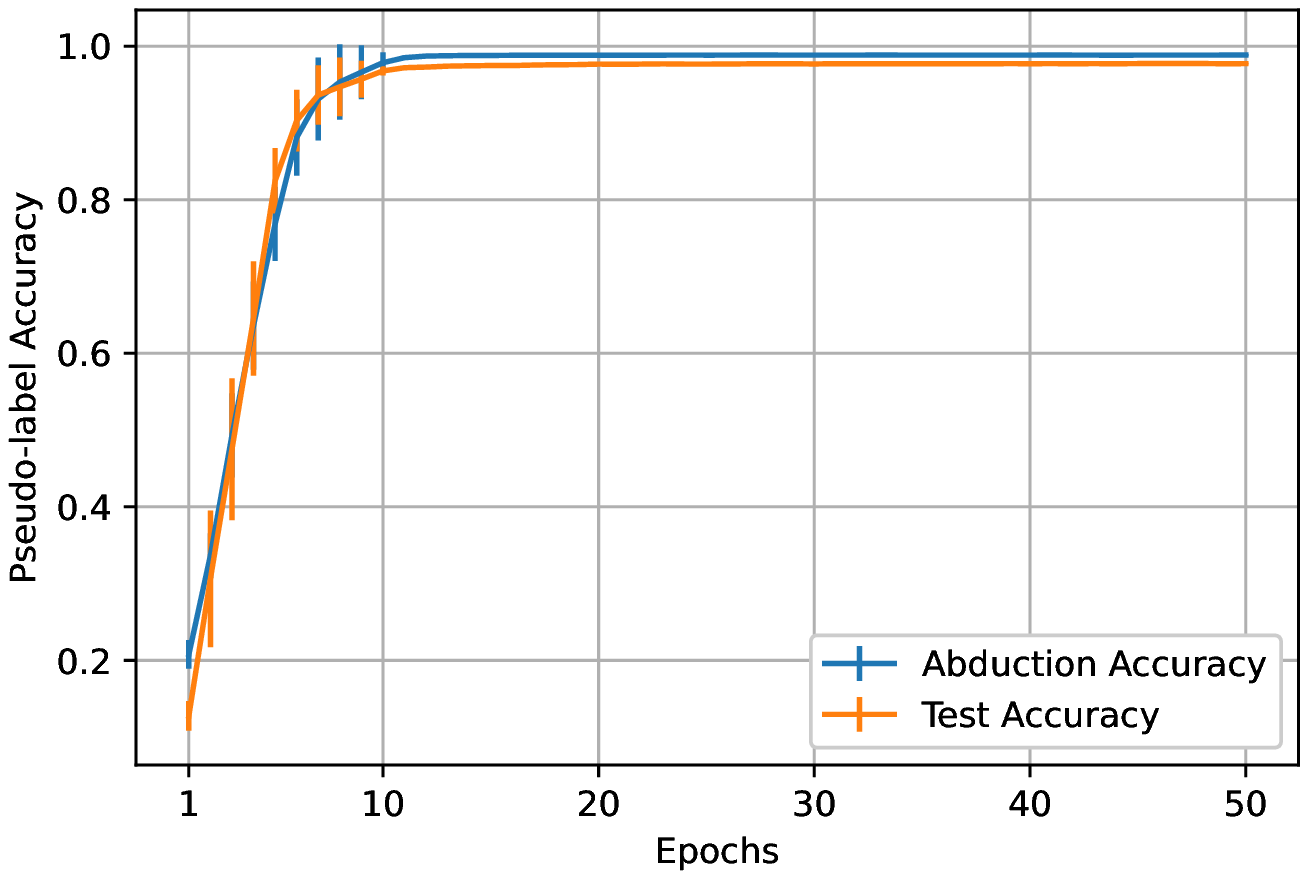}
    \caption{MNIST product}
    \end{subfigure}\\
    \centering
    \begin{subfigure}[b]{0.45\textwidth}
    \includegraphics[width=\textwidth]{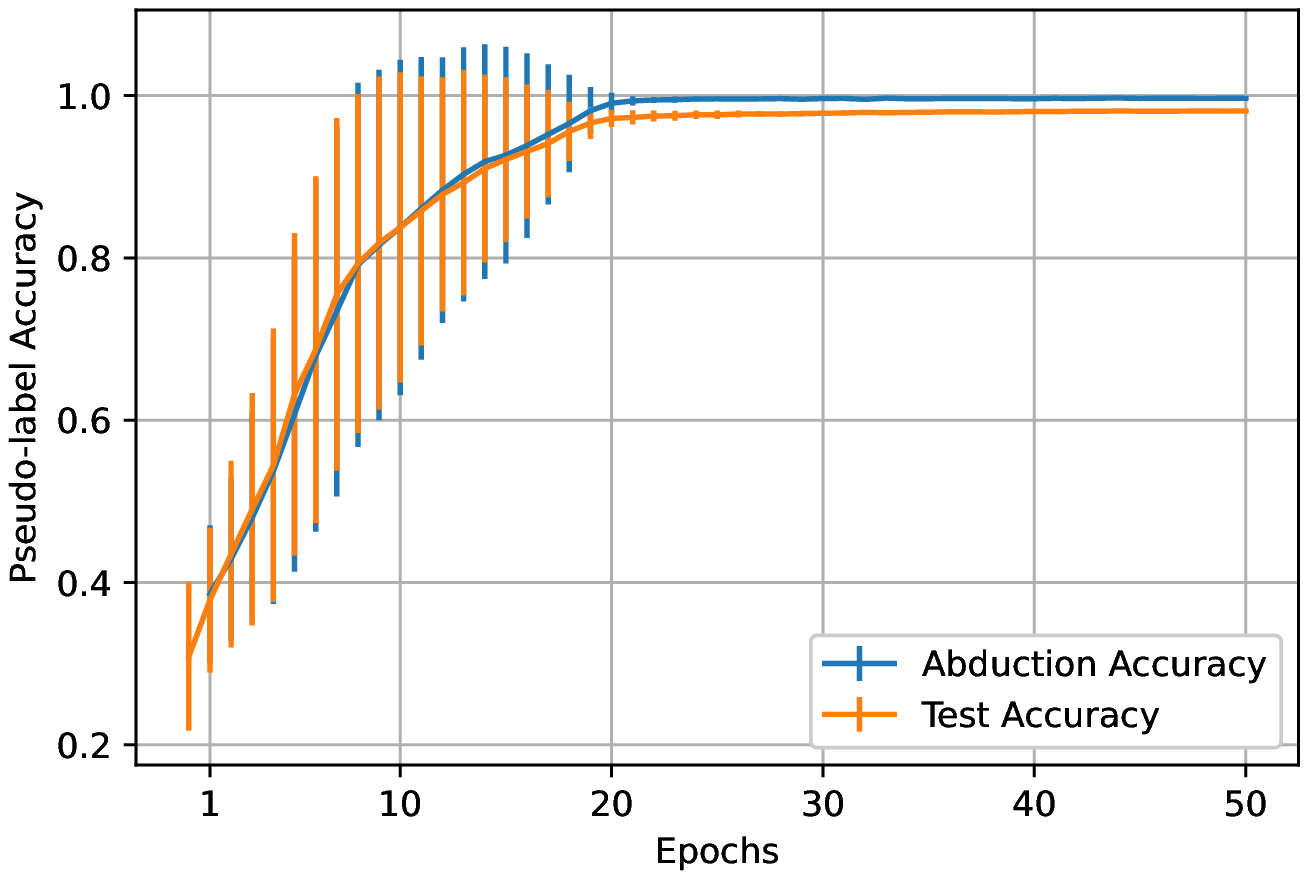}
    \caption{MNIST sum with 1-shot CNN pre-train}
    \end{subfigure}
    ~
    \begin{subfigure}[b]{0.45\textwidth}
    \includegraphics[width=\textwidth]{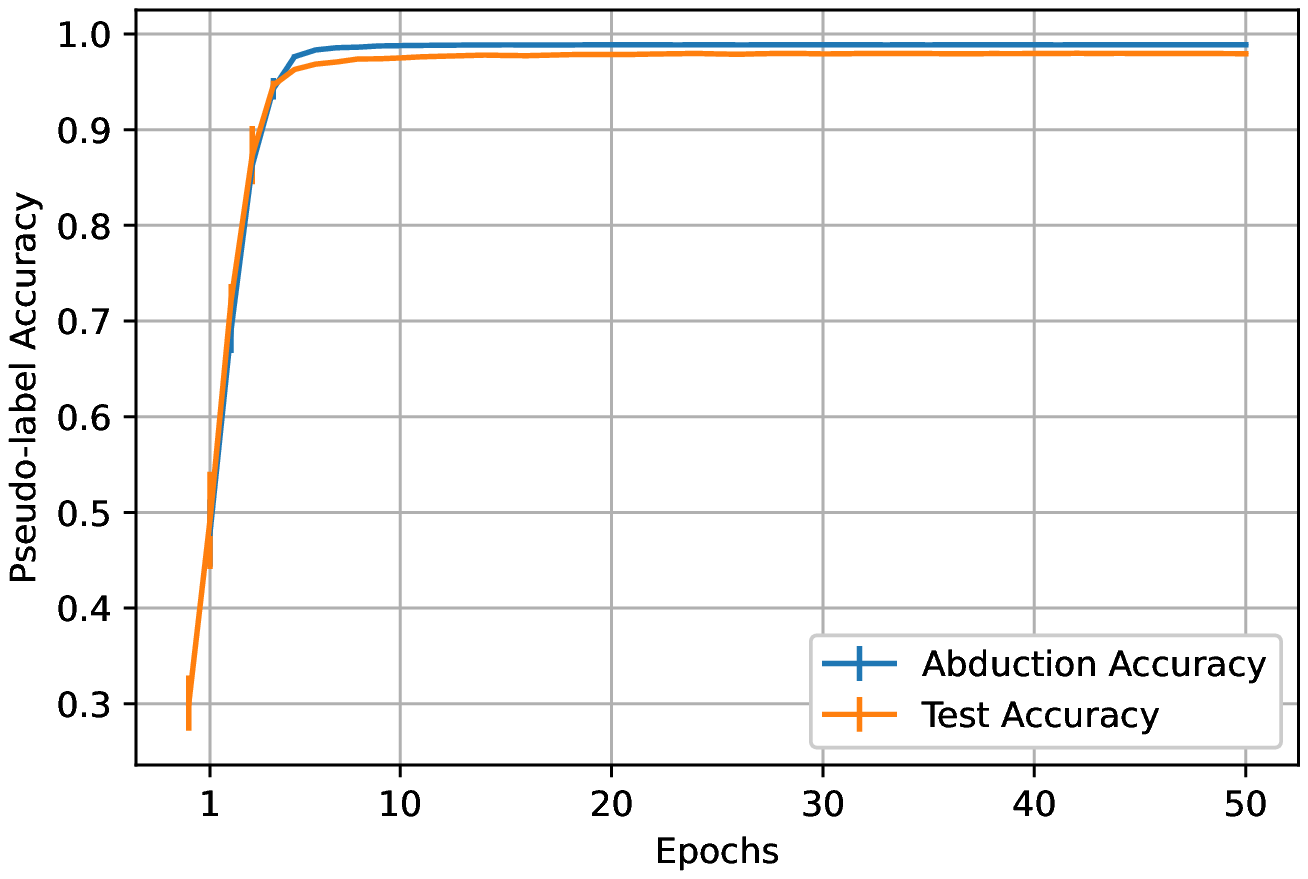}
    \caption{MNIST product with 1-shot CNN pre-train}
    \end{subfigure}
    \caption{Pseudo-label accurracy during $Meta_{Abd}$ and $Meta_{Abd+\text{1-shot CNN}}$ learning.\label{fig:acc_exp1}}
\end{figure*}

\begin{figure*}[!t]
    \centering
    \fbox{
    \begin{minipage}{0.6\linewidth}
        \texttt{nn(\img{figs/1.png},0,P00). nn(\img{figs/1.png},1,P01). nn(\img{figs/1.png},2,P02). ...}\\
        \texttt{nn(\img{figs/2.png},0,P10). nn(\img{figs/2.png},1,P11). nn(\img{figs/2.png},2,P12). ...}\\
        \texttt{nn(\img{figs/3.png},0,P20). nn(\img{figs/3.png},1,P21). nn(\img{figs/3.png},2,P22). ...}\\
        \texttt{nn(\img{figs/9.png},0,P30). nn(\img{figs/9.png},1,P31). nn(\img{figs/9.png},2,P32). ...}
    \end{minipage}
    }%
    \caption{Monadic probabilistic facts generated by neural network in the sum/product tasks.\label{fig:facts1}}
\end{figure*}

\begin{figure}[t!]
  \centering
  \resizebox{0.48\textwidth}{!}{
    \begin{tabular}{|l|}
      \hline
      \texttt{\it \% List operations.}\\
      \texttt{head([H|\_],H).}\\
      \texttt{tail([\_|T],T).}\\
      \texttt{empty([]).}\\
      \\
      \texttt{\it \% Background knowledge about permutation}\\
      \texttt{permute(L1,O,L2):-}\\
      \texttt{\hspace{2em}length(L1,N),}\\
      \texttt{\hspace{2em}findall(S,between(1,N,S),O1),}\\
      \texttt{\hspace{2em}\it \% generate permutation with Prolog's built-in predicate}\\
      \texttt{\hspace{2em}catch({\color{red}permutation}(O1,O),\_,fail),}\\
      \texttt{\hspace{2em}permute1(L1,O,L2).}\\
      \texttt{\it \% permute the image list with order O}\\
      \texttt{permute1([],[],\_).}\\
      \texttt{permute1([S|List],[O|Os],List2):-}\\
      \texttt{\hspace{2em}nth1(O,List2,S),}\\
      \texttt{\hspace{2em}permute1(List,Os,List2).}\\
      \\
      \texttt{\it\% {\color{red}Abducible} primitives.}\\
      \texttt{abduce\_{\color{red}nn\_pred}([X,Y|\_],{\color{blue}nn\_pred(X,Y)},Score):-}\\
      \texttt{\hspace{2em}{\color{red}nn\_pred}(X,Y,Score).}\\
      \hline
    \end{tabular}%
  }
  \caption{Background knowledge used in the MNIST sorting task.\label{fig:BK2}}
\end{figure}
  
We also report the pseudo-label accuracy of abduction and perception during training, which are shown
in Fig.~\ref{fig:acc_exp1}. The blue lines are the accuracy of the abduced labels (i.e., the accuracy
of the expectation of $z$) in each EM iteration; the orange lines are the accuracy of the perceptual
neural net's classification accuracy on the MNIST test set. As we can observe, the convergence
speed of cumulative sum is slower, because its the posterior 
distribution on pseudo-labels ($P(H,z|B,x,y,\theta)$) is much denser than that of 
cumulative product. After applying the one-shot CNN pre-train, whose test accuracy is shown at 0 epoch
in the figures, the convergence speed of MNIST cumulative sum is significantly improved because the EM
algorithm is less-likely to be trapped in local optimums.

\subsection{MNIST Sorting}
\label{seq:impl_srt}

Different to the MNIST cumulative sum/product tasks which learn a perceptual neural network predicting
the digit in each single image, in the MNIST sorting task, $Meta_{Abd}$ uses a perceptual neural network
to learn an unknown binary relation between two images. Examples are shown in Fig.~\ref{fig:facts2}. 
The neural network uses the same convnet as before to take the input from a pair of images (the first
two arguments of predicate \texttt{nn}), and then a Multi-Layered Perception (MLP) is used to predict 
the probability \texttt{PIJ}. The first two clauses translate the neural network's output \texttt{nn}
to the probabilistic facts for $Meta_{Abd}$'s abduction.

\begin{figure*}[!t]
  \centering
  \fbox{
    \begin{minipage}{0.75\linewidth}
      \texttt{nn\_pred(X,Y,P) :- nn(X,Y,P), !.}\\
      \texttt{nn\_pred(X,Y,P) :- nn(Y,X,P1), P is 1-P1, !.}\\\\
      \texttt{nn(\img{figs/1.png},\img{figs/2.png},P01). nn(\img{figs/1.png},\img{figs/3.png},P02). nn(\img{figs/1.png},\img{figs/9.png},P02). ...}\\
      \texttt{nn(\img{figs/2.png},\img{figs/3.png},P12). nn(\img{figs/2.png},\img{figs/9.png},P13). nn(\img{figs/3.png},\img{figs/9.png},P13). ...}
    \end{minipage}
  }%
  \caption{Dyadic probabilistic facts generated by neural network in the sorting task.\label{fig:facts2}}
\end{figure*}

\begin{figure*}[!t]
  \centering
  \includegraphics[width=0.5\textwidth]{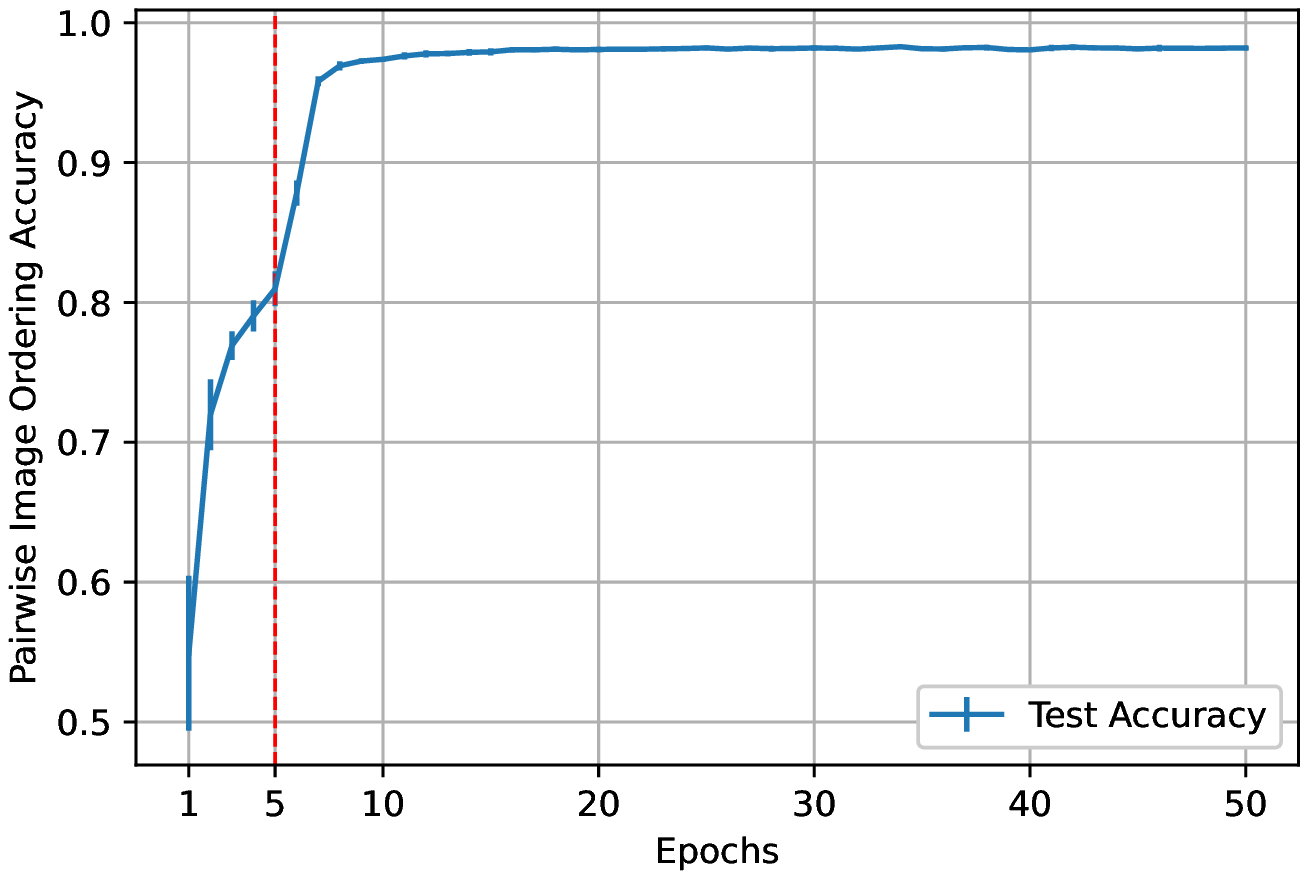}
  \caption{MNIST pairwise ordering (\texttt{nn\_pred}) accuracy during learning.\label{fig:acc_exp2}}
\end{figure*}

\paragraph{Example (Dyadic facts abduction)}
Background knowledge of the MNIST sorting task is shown in Fig.~\ref{fig:BK2}. 
Different to the previous example which abduces the label of each input image, in the sorting task,
the facts being extracted from raw data are dyadic relationship between two images. Given an training
example with input $x=\texttt{[\img{figs/1.png},\img{figs/2.png},\img{figs/3.png},\img{figs/9.png}]}$,
the perceptual neural network will process all the pairwise combinations among them and output a score
as shown in Fig.~\ref{fig:facts2}. Because the pairwise combinations are just a half of pairwise 
permutations, we also provided a symmetric rule to complete them (the first two clauses in 
Fig.~\ref{fig:facts2}). During $Meta_{Abd}$'s induction, the abduced facts are the pairwise
probabilistic facts themselves instead of CLP(Z) constraints like before, so the \texttt{Score}
is the probability of each probabilistic fact. In other words, in the sorting task,
\emph{the abduction of $z$} (the truth values of the probabilistic facts) \emph{is performed simultaneously with
logical induction}. Recall the Prolog code of $Meta_{Abd}$ in Fig.~\ref{fig:meta_abd}, there
is a greedy process that keeps the current most probable abduction with \texttt{getmaxprob(Max)}
and \texttt{setmaxprob(Max)}. The greedy strategy is used to prune the search space of $z$, 
it excludes the facts with low probability and quickly find a locally optimal $z$ (truth value
assignment), which will be used as pseudo-labels to train the perceptual neural network
in the maximisation step.

\begin{figure*}[!t]
    \centering
    \fbox{
    \begin{minipage}{0.6\linewidth}
    \small
        \texttt{metarule([P,Q],[P,A],[[Q,A]]).\\
metarule([P,Q],[P,A],[[Q,A,B],[P,B]]).\\
metarule([P,Q,R],[P,A],[[Q,A,B],[R,B]]).\\
metarule([P,Q,R],[P,A,B],[[Q,A],[R,A,B]]).\\
metarule([P,Q],[P,A,B],[[Q,A,B]]).\\
metarule([P,Q,R],[P,A,B],[[Q,A,B],[R,A,B]]).\\
metarule([P,Q,R],[P,A,B],[[Q,A,B,C],[R,C]]).\\
metarule([P,Q,R],[P,A,B],[[Q,A,B],[R,B]]).\\
metarule([P,Q,R],[P,A,B],[[Q,A,C],[R,C,B]]).}
    \end{minipage}
    }%
    \caption{Meta-rules used in all the experiments.\label{fig:meta-rule}}
\end{figure*}

Fig.~\ref{fig:acc_exp2} shows the perception accuracy during training. The test pairs contains
10,000 randomly sampled images from the MNIST test set. The vertical line at epoch 5 shows the
time point when $Meta_{Abd}$ switching from the sub-task (learning concept of ``sorted'' 
with target predicate \texttt{s}) to the main tasks (learning permutation sort). The results in
this figure verifies that the perception model is successfully re-used in this experiment.

\subsection{Reproducibility}

We introduce more experimental details in this subsection. All experiments are completed on a PC with
AMD Ryzen 3900X CPU and Nvidia 2080Ti GPU. The data and source codes of $Meta_{Abd}$ will be available
after the publication of this work.

\subsubsection{meta-rules}

The meta-interpreter of $Meta_{Abd}$ uses a set of meta-rules to guide the induction of the logic theory $H$. We use the 
meta-rules from the higher-order meta-interpreter $Metagol_{ho}$\footnote{https://github.com/andrewcropper/mlj19-metaho}~\cite{cropper:20:metaho},
which are shown in Fig.~\ref{fig:meta-rule}.It has been shown that these
meta-rules have universal Turing expressivity and can represent higher-order
programs~\cite{cropper:20:metaho}.

We further compared the inference speed of $Meta_{Abd}$ with different sizes of
meta-rules. Specifically, following are the time difference measured by the
average number of Prolog inferences in each batch of 's abduction-induction
inference in the accumulative sum task. The settings are as follows:

\begin{itemize}
\item $Meta_{Abd}$ contains at least one metarule, which is \texttt{P(A,B):-Q(A,B)},
  i.e., calling a primitive function. However, it is not complete for
  representing the hypothesis space since none of the primitive predicates is
  able to define the target concept (otherwise they won't be called as
  ``primitives''). Hence, we start from at least 2 meta-rules;
\item The perceptual CNN is randomly initialised and un-trained, i.e., the
  distribution of probabilistic facts is random, which is the \emph{worst-case} for
  abduction, so the result here is slower than the average result in Fig.~\ref{fig:time};
\item Choosing metarules is a subset selection problem. Following the
  traditions in combinatorial optimisation, we report the worst result among
  all varied combinations;
\item The number of Prolog inferences includes the CLP(Z) optimisation.
\end{itemize}
 
\begin{table}
  \centering
  \begin{tabular}{ccc}
    \toprule
    \textbf{Number of meta-rules} & \textbf{Number of Prolog inferences} &
                                                                           \textbf{Time
                                                                           (seconds)}
    \\
    \midrule
    9 & 26324856 & 1.571 \\
    8 & 26324638 & 1.567 \\
    7 & 26324626 & 1.567 \\
    6 & 26324287 & 1.527 \\
    5 & 26324009 & 1.528 \\
    4 & 26321479 & 1.521 \\
    3 & 26314047 & 1.521 \\
    2 & 10991735 & 0.635 \\
    \bottomrule
  \end{tabular}
  \caption{Time costs (worst-case) of using different numbers of meta-rules.
    Note that the setting of using only 2 meta-rules is equivalent to RNNs
    which are forced to learn a minimum recursive program.\label{fig:mr_time}}
\end{table}

As we can see from Fig.~\ref{fig:mr_time}, there is not much difference between
using 9--3 metarules for $Meta_{Abd}$ (when the program hypothesis space is
complete). Hence, if the users have a strong bias on the target theory and only use the relevant
metarules, the search speed can be very fast.

\subsubsection{Neural Network \& Hyperparameters}

The convnet in our experiments is from PyTorch's MNIST
tutorial\footnote{https://github.com/pytorch/examples/tree/master/mnist}
as~\cite{nalu18} suggested. The LSTM and RNN models in the MNIST cumulative
sum/product experiments have 2 hidden layers with dimension 64; the \texttt{NAC}
and \texttt{NALU} modules have 2 hidden layers with dimension 32. In the MNIST
sorting experiments, we set the hyperparameter $\tau=1.0$ for NeuralSort, which is
the default value in the original
codes\footnote{https://github.com/ermongroup/neuralsort}. Moreover, the output
of NeuralSort is a vector with floating numbers, in order to reproduce the
result from the original paper, we rank the output scores to generate the final
prediction of orderings.

DeepProblog~\cite{manhaeve:18:deepproblog} and
Neural Logical Machines (NLM)~\cite{dong2018neural} are treated as blackbox models
attached with the same convnet as $Meta_{Abd}$. The Problog programs of
DeepProblog are the ground-truth programs in Fig.~\ref{fig:prog}; the parameters
of NLM is tuned following the instructions in its repository.
\end{document}